\pdfoutput=1

\documentclass[11pt]{article}

\usepackage{coling}

\usepackage{times}
\usepackage{latexsym}

\usepackage{amsmath,amsfonts,amssymb,amsthm}
\usepackage{multirow,multicol}
\usepackage{booktabs}
\usepackage{float}
\usepackage{longtable}
\usepackage{paralist}
\usepackage{numprint}
\usepackage{latexsym}
\usepackage{graphicx}
\usepackage{caption}
\usepackage{subcaption}
\usepackage{url}
\usepackage{tikz}
\usepackage{pgfplots}

\usepackage{amsmath,amsfonts,amssymb,amsthm}
\usepackage{tikz}
\usepackage{pgfplots}
\usepackage{pgfmath}
\usetikzlibrary{plotmarks,shapes,pgfplots.dateplot}
\usetikzlibrary{math,calc,trees,positioning,arrows,chains,shapes.geometric,
decorations.pathreplacing,decorations.pathmorphing,shapes,
matrix,shapes.symbols}
\usetikzlibrary{3d,decorations.text,shapes.arrows,fit,backgrounds,quotes,arrows.meta}
\usepackage{ifthen}
\usepackage{multirow,multicol}
\usepackage{booktabs}
\usepackage{float}
\usepackage{longtable}
\usepackage{latexsym}
\usepackage{graphicx}
\usepackage{caption}
\usepackage{subcaption}
\usepackage{url}

\colorlet{color01}{red!10!white}
\colorlet{color02}{orange!10!white}
\colorlet{color03}{green!10!white}
\colorlet{color04}{blue!10!white}
\colorlet{color05}{magenta!20!white}
\colorlet{color06}{black!15!white}
\colorlet{color07}{yellow!10!white}
\colorlet{color08}{white}
\colorlet{color1}{red!80!black}
\colorlet{color1a}{red!50!white}
\colorlet{color2}{orange!80!black}
\colorlet{color2a}{orange!50!white}
\colorlet{color3}{green!50!black}
\colorlet{color3a}{green!90!black}
\colorlet{color4}{blue!80!black}
\colorlet{color4a}{blue!50!white}
\colorlet{color5}{magenta!80!black}
\colorlet{color5a}{magenta!50!white}
\colorlet{color6}{black!80!black}
\colorlet{color6a}{black!50!white}
\colorlet{color7}{white!30!black}
\colorlet{color7a}{black!50!white}
\colorlet{color8}{yellow}
\colorlet{color8a}{black!50!yellow}

\tikzset{every picture/.style={semithick},every path/.style={thick,rounded corners,->}}
\tikzset{
token/.style={rectangle,rounded corners,draw=black,thick,inner sep=4pt,outer sep=0,minimum width=2em,fill=color03,text centered, minimum height=1.5em,font=\small\sffamily},
tokenperm/.style={token,fill=color04},
tokenres/.style={token,fill=color02},
numperm/.style={token,fill=color08},
ne/.style={ draw=none, fill=none, font=\footnotesize\sffamily,  minimum height=0em, text centered},
ncirc/.style={ circle, draw=black, thin, fill=none, font=\footnotesize\sffamily,  minimum height=2em, inner sep=0, text centered},
nd/.style={ font=\sffamily,  text centered},
nodecomp/.style={ rectangle,  rounded corners,  draw=black, thick, text width=2em,  font=\footnotesize\sffamily, minimum height=1.3em,  text centered},
nodevar/.style={ nodecomp, fill=green!10,
},
diablo/.style={ rectangle,  rounded corners,  draw=black, thick, text width=10em,  font=\footnotesize\sffamily, minimum height=1em,  text centered},
branch/.style ={circle,inner sep=0pt,minimum size=1.5mm,fill=black,draw=black},
diablo2/.style={ rectangle,  rounded corners,  fill=red!10, draw=black!80,thick, text width=3em,  font=\footnotesize\sffamily, minimum height=2.7em,  text centered},
diafeature/.style={ rectangle, rounded corners=2pt,  fill=green!10, draw=black!80,thick, text width=1em,  font=\footnotesize\sffamily, minimum height=6em,  text centered},
diafeatnarr/.style={ rectangle, rounded corners=2pt,  fill=green!10, draw=black!80,thick, text width=.5em,  font=\footnotesize\sffamily, minimum height=6em,  text centered},
dialoss/.style={ diablo2, fill=green!10,
},
rotnode/.style={ anchor=center, rotate=90, font=\footnotesize\sffamily
},
diaext/.style={ diablo2,fill=yellow!40, },
diablo3/.style={rectangle, rounded corners, fill=blue!10, draw=blue!40,thick, text width=3.5em,  font=\footnotesize\sffamily\bfseries, text=blue, minimum height=1.5em, text centered},
line/.style={draw=red,rounded corners,thick, ->, decoration={markings,mark=at position 1 with {\arrow[scale=4,>=stealth]{>}}},postaction={decorate}},
element/.style={ tape, top color=white, bottom color=blue!50!black!60!, minimum width=8em, draw=blue!40!black!90, very thick, text width=10em, minimum height=3.5em, text centered, on chain},
every join/.style={->,rounded corners,thick,shorten >=1pt},  decoration={brace},
lineblue/.style={    join,line width=.07cm,->,blue!20  }
}

\def\RR{\mathbb{R}}
\def\x{\mathbf{x}}

\def\btheta{\boldsymbol{\theta}}

\newcommand{\brc}[2]{\left(#1\middle| #2\right)}

\newcommand{\pcth}[2]{p_{\btheta}\brc{#1}{#2}}

\newcommand\EEEE[2]{{\mathbb E}_{#1}\left[ #2 \right]}

\newcommand{\myexp}[1]{e^{#1}}

\DeclareFontFamily{U}{mathx}{\hyphenchar\font45}
\DeclareFontShape{U}{mathx}{m}{n}{<-> mathx10}{}
\DeclareSymbolFont{mathx}{U}{mathx}{m}{n}
\DeclareMathAccent{\widebar}{0}{mathx}{"73}

\newcommand{\z}{\ensuremath{\boldsymbol{z}}}

\usepackage{mathtools}

\definecolor{light-gray}{gray}{0.80}

\definecolor{darkred}{rgb}{0.64, 0.0, 0.0}

\theoremstyle{definition}

\newtheorem*{thm*}{Theorem}

\DeclareMathOperator*{\argmax}{arg\,max}

\def\RR{\mathbb{R}}
\def\x{\mathbf{x}}
\def\z{\mathbf{z}}

\def\btheta{\boldsymbol{\theta}}

\def\btheta{\boldsymbol{\theta}}

\def\sth{s_{\btheta}}

\newcommand{\picth}[2]{\pi_{\btheta}\brc{#1}{#2}}

\def\csn{\emph{CodeSearchNet}\xspace}
\def\gh{\emph{GitHub}\xspace}
\def\poj{{POJ-104}\xspace}
\def\xcd{{XCD}\xspace}
\def\advtest{\emph{AdvTest}\xspace}

\usepackage{algorithm2e}

\SetCommentSty{mycommfont}
\SetKwComment{Comment}{/* }{ */}
\RestyleAlgo{ruled}

\def\cbert{\ensuremath{\text{CodeBERT}}\xspace}
\def\cbertb{\ensuremath{\text{CodeBERT}_{\mathrm{base}}}\xspace}
\def\gcbert{\ensuremath{\text{GraphCodeBERT}}\xspace}
\def\gcbertb{\ensuremath{\text{GraphCodeBERT}_{\mathrm{base}}}\xspace}

\def\gcbertbp{\ensuremath{\text{GraphCodeBERT}_{\mathrm{base}}^{\mathrm{POJ}}}\xspace}
\def\cct{\ensuremath{\text{CCT}}\xspace}
\def\cctlm{\ensuremath{\text{CCT-LM}}\xspace}

\begin{document}

\title{CCT-Code: Cross-Consistency Training for Multilingual Clone Detection and Code Search}

\author{
\textbf{Anton Tikhonov\textsuperscript{1}},
\textbf{Nikita Sorokin\textsuperscript{1}},
\textbf{Dmitry Abulkhanov\textsuperscript{1}},\\
\textbf{Irina Piontkovskaya\textsuperscript{2}},
\textbf{Sergey Nikolenko\textsuperscript{3}},
\textbf{Valentin Malykh\textsuperscript{1}},
\\
\textsuperscript{1}MTS AI,
\textsuperscript{2}Huawei Noah’s Ark Lab,\\
\textsuperscript{3}St. Petersburg Department of the Steklov Institute of Mathematics
\\
\small{
\textbf{Correspondence:} \href{mailto:valentin.malykh@phystech.edu}{valentin.malykh@phystech.edu}
}
}

\maketitle

\begin{abstract}

We consider the well-known and important tasks of clone detection and information retrieval for source code. The most standard setup is to search clones inside the same language code snippets. But it is also useful to find code snippets with identical behaviour in different programming languages.  Nevertheless multi- and cross-lingual clone detection has been little studied in literature.  We present a novel training procedure, cross-consistency training (CCT) leveraging cross-lingual similarity, that we apply to train language models on source code in various programming languages. We show that this training is effective both for encoder- and decoder-based models.
The trained encoder-based CCT-LM model
achieves a new state of the art on POJ-104 (monolingual C++ clone detection benchmark) with 96.73\% MAP and AdvTest (monolingual Python code search benchmark) with 47.18\% MRR. The decoder-based CCT-LM model shows comparable performance in these tasks. In addition, we formulate the multi- and cross-lingual clone detection problem and present XCD, a new benchmark dataset produced from CodeForces submissions.
\end{abstract}

\section{Introduction}

The clone detection problem stems from an important need arising in software development practice: to identify code with the same behaviour and effective output; this is useful, e.g., for code unification, refactoring and control of side effects.
\citet{POJ} formulated the clone detection task for {C/C++} source code, and later the problem was extended to other programming languages. The natural next step for this problem is to detect the same behaviour for code across different programming languages.\footnote{For an illustrative example, we refer here to the \url{http://helloworldcollection.de/} website that contains ``Hello, world!'' snippets in 603 programming languages.} In this work, we formulate the multilingual clone detection task, collect a dataset, and establish reasonable baselines.

Various approaches for the clone detection task have been developed, starting from algorithmic methods~\cite{baker1993program,krinke2001} and continuing with machine learning models~\cite{li2017cclearner,thaller2020towards,gotmare2021cascaded}. Most machine learning approaches are based on learning the representations (embeddings) of code snippets; this approach allows to find duplicate code snippets by similarity between their embeddings in the latent space. The performance of such systems critically depends on the quality of embeddings. In this work, we present a novel training technique and pipeline called \cct for language models that allows them to embed code snippets in an efficient way. We demonstrate this by achieving state of the art on a preexisting clone detection dataset POJ-104~\cite{POJ} and on the newly formulated multilingual clone detection dataset \xcd.
Interestingly, we also show that the \cct technique allows a model to produce representations useful for code search as well. Code search, as formulated by \citet{lu_codexglue_2021}, is the task where the query is a text description and the possible documents are code snippets.

Thus, the main contributions of our work are:
\begin{inparaenum}[(1)]
\item a pretraining method \cct that allows a model to align code snippets in different programming languages;
\item a novel multilingual clone detection dataset \xcd based on \emph{CodeForces} submissions;
\item new state of the art results obtained by \cctlm model trained with \cct on the clone detection datasets POJ-104 and \xcd\footnote{We will release the \xcd dataset and the source code for \cct and \cctlm upon acceptance.};
\item state of the art results of \cctlm for code search on the \advtest dataset.
\end{inparaenum}
The paper is organized as follows: Section~\ref{sec:rel_work} surveys related work,
Section~\ref{sec:data} discusses code search datasets and introduces \xcd, Section~\ref{sec:method} shows the \cct pretraining approach, Section~\ref{sec:eval} presents our results,
Section~\ref{sec:analysis} analyzes them and presents an ablation study, Section~\ref{sec:concl} concludes the paper, and Section~\ref{sec:limits} discusses the limitations of our work.

\section{Related Work}
\label{sec:rel_work}

Our methods are inspired by natural language processing, thus related work includes both pure NLP and source code processing.

\textbf{Datasets}.
\citet{husain2019codesearchnet} presented the \csn dataset constructed from a \emph{GitHub} dump where the authors split method bodies into the code itself and a description. This dataset contains 2 million code snippet-description pairs in 6 programming languages, including \emph{Python}.
This dataset was partially used by \citet{hasan2021codesc} who combined \csn and three other datasets into a larger one. From \csn they used the \emph{Java} part and \emph{Python} part translated automatically into \emph{Java}. The resulting dataset contains 4 million code snippet-description pairs.
\emph{CodeXGLUE} presented by \citet{lu_codexglue_2021} is a machine learning benchmark collection of datasets for code understanding and generation tasks, which includes a modification of \csn. \emph{CodeXGLUE} provides a benchmark for various code-to-code, code-to-text, text-to-code tasks, including code search. This benchmark includes code in 10 programming languages.
There are two main datasets for clone detection: \poj~\cite{POJ} and \emph{BigCloneBench}~\cite{wang2020detecting}. \poj represents a comparatively small corpus of C++ solutions from a student judging system. \emph{BigCloneBench} comprises a vast dataset containing automatically mined data in the Java language.

\textbf{Code Search}.
Early works on code search mostly relied on classic information retrieval~\cite{bacchelli2010linking,brandt2010example,campbell2017nlp2code,chan2012searching}.
\citet{brandt2010example}, \citet{barzilay2013facilitating}, and \citet{ponzanelli2014mining} utilized existing industrial Web search engines.
\citet{gu2018deep} used modern dense vector representations for information retrieval, training
two recurrent neural networks to represent source code and text respectively. \citet{feng_codebert_2020} presented a language model-based approach to produce these representations. \citet{gotmare2021cascaded} used three Transformer-based models, two as encoders and one as a classifier, for a hierarchical representation of code and text; although they experimented with sharing encoder parameters, it lowered the final quality of their model. Our model, in contrast, uses a single decoder part of the transformer to embed queries and documents and skips the classifier part.

\textbf{Clone Detection}.
To the best of our knowledge, the attempt to detect similar program code was made by Baker~\cite{baker1993program} who proposed to find duplicate code in exact and parameterized forms. The latter approach could be reformulated as code obfuscation and search for an obfuscated exact match. \citet{krinke2001} proposed to apply graph theory, namely to compare shortest paths of abstract syntactic trees extracted from the code snippets. Early work on algorithmic clone detection was summarized by \citet{roy2007}. Last but not least, we note one of the most recent algorithmic approaches \emph{SourcererCC}~\cite{sajnani2016sourcerercc}, which proposes a sophisticated algorithm that includes the construction of a search index for candidate clone snippets.
Recent approaches are usually based on machine learning. Probabilistic approaches include \citet{alomari2018towards} with Latent Dirichlet allocation and \citet{thaller2020towards} with a novel graph probabilistic model based on density estimation.
One of the first successful deep learning approaches was CClearner~\cite{li2017cclearner} that used text extracted from a program and its AST features and had a simplistic multilayer perceptron architecture for clone classification on a closed code base. More recent deep learning models include graph neural networks on ASTs~\cite{wang2020detecting} and employ pretrained language models~\citet{AddLeak}.

\textbf{Language models for source code}.
After the success of BERT-like models for natural language, these approaches were also applied to programming languages. Several pre-trained programming language models were presented recently, including: \cbert~\cite{feng_codebert_2020}, which is a bimodal pre-trained model for source code and natural language based on the RoBERTa Transformer architecture~\cite{RoBERTa}, trained on masked language modeling (MLM) and replaced token detection objectives;
\gcbert~\cite{guo_graphcodebert_2021} that uses data flow during pre-training to solve MLM, edge prediction, and node alignment tasks; SynCoBERT~\cite{wang_syncobert_2021} that uses multimodal contrastive learning and is pre-trained on identifier prediction and AST edge prediction.
In recent times, autoregressive decoder models like DeepSeek-Coder~\cite{deepseekcoder} have taken over the world. They represent the decoder part of the transformer model, which is pre-trained on a huge corpus of source code. They are used in source code generation tasks, such as code completion, documentation generation, and so on.

\textbf{Multilingual NLP}.
Open-domain question answering task assumes answering factoid questions without a predefined domain~\cite{kwiatkowski2019natural}.
Since we propose a novel benchmark in cross-lingual code understanding, we also review multilingual NLP models, where
recent research has been focused on non-English question answering (QA) datasets and multilingual transfer learning, usually from English to other languages.

Until recently, appropriate training and evaluation datasets have been scarce, but in recent years many works have collected loosely aligned data obtained through automatic translation or by parsing similar multilingual sources.
Thus,
\citet{ahn2004cross} presented an early attempt to multilingual question answering by translating an answer to the target language.
Two years later,
\citet{bos2006cross} presented another similar system using translation.
Lee and Lee~\cite{lee2019cross} showed successful transfer for multilingual QA with training on English data and evaluation in Chinese.
\citet{ferrandez2007applying} presented a work on multilingual knowledge base incorporating a system.
\citet{mitamura2006keyword} developed a system with the translation of keywords extracted from question to get an answer in the target language.
\citet{artetxe-etal-2020-cross} studied multilingual transfer of monolingual representations for a Transformer-based masked language model.
\citet{mhamdi-etal-2021-x} examined a multilingual optimization-based meta-learning approach to learn to adapt to new languages for QA.
\citet{cocondeser} proposed unsupervised pretraining for dense passage retrieval, although the authors concentrated on retrieval itself and ignored the multilingual nature of the data.
Most approaches used extractive models, mostly due to the prevalence of datasets where, similar to SQuAD~\cite{rajpurkar2016squad}, labeled data includes an explicitly stated question, a passage containing an answer, and a span labeling for the answer. However, several works have considered generative QA:
\citet{kumar-etal-2019-cross} and
\citet{chi2019crosslingual} studied multilingual question generation,
\citet{riabi2020synth} suggested a method to produce synthetic questions in a multilingual way by using multilingual MiniLM, and
\citet{shakeri2020multi} proposed a method to generate multilingual question-answer pairs by a generative model (fine-tuned multilingual T5) based on samples automatically translated from English to the target language.
Generative QA was mostly considered for datasets with long answers, but the
generative model FiD~\cite{fid} achieved competitive results on SQuAD-like datasets, where an answer is supposed to be a short text span. For open domain QA,
\citet{rag} used generative models in their RAG approach that processes top $k$ passages from the retriever in the encoder simultaneously and uses their representations in the decoder for answer generation, in a process called fusion.
Processing the passages independently in the encoder allows a model to scale to many contexts, as it only runs self-attention over one context at a time.
The FiD model follows this paradigm, further improving the results in question generation.
For QA over a knowledge graph,
\citet{zhou-etal-2021-improving} studied unsupervised bilingual lexicon induction for zero-shot multilingual transfer for multilingual QA in order to map training questions in the source language into questions in the target language for use as augmented training data, which is important for zero-resource languages.

\section{Datasets}\label{sec:data}

In this work we use two kinds of datasets, one for clone detection and another for code search.

\textbf{Code Search}.
For code search we use the \csn dataset introduced by~\citet{husain2019codesearchnet}. The original version of \csn consists of natural language queries paired with most relevant code snippets in six programming languages. Each snippet represents the code of a function collected from \gh open source code.

\textbf{CodeSearchNet AdvTest} is a Python-only dataset constructed from the \csn corpus by
\citet{lu_codexglue_2021}. Each example, again, includes a function paired with a text document, and similar to
\citet{husain2019codesearchnet}, \advtest takes the first paragraph of the documentation as the query.
\citet{lu_codexglue_2021} make an interesting observation: after normalizing function and variable names with special tokens, the Mean Reciprocal Rank (MRR) scores of RoBERTa~\cite{RoBERTa} and CodeBERT~\cite{feng_codebert_2020} models for code search on the original \csn dataset drop from 0.809 to 0.419 and from 0.869 to 0.507 respectively (in \emph{Python}).
So, to improve the quality of the dataset and make it harder, they first filtered the data by removing examples for which
the code could not be parsed into an abstract syntax tree,
the document was shorter than 3 or longer than 256 tokens,
the document contained special tokens such as ``http://'', or
the document was empty or not written in English.
The filtered dataset contains \numprint{251820} / \numprint{9604} / \numprint{19210} examples in its training/validation/test sets respectively.

Then, to better test the understanding and generalization abilities of the model,
\citet{lu_codexglue_2021} normalized function and variable names in testing and development sets to nondescript tokens such as \emph{func} for the function name and \emph{arg\_i} for the $i$-th variable name.
The task remains to search for source code based on a natural language query.
Moreover, in contrast to the testing phase of previous works~\cite{husain2019codesearchnet,feng_codebert_2020} that only involved \numprint{1000} candidates,
\citet{lu_codexglue_2021} used the entire test set for each query, which makes the \advtest dataset even more difficult. The training set comes from the filtered CodeSearchNet dataset~\cite{husain2019codesearchnet} where the code is represented in a raw form in addition to tokenization native to its programming language.
\advtest uses MRR as the basic evaluation metric, defined as
$\mathrm{MRR@Q} = \frac{1}{Q}\sum_{i=1}^{Q}\frac{1}{\mathrm{rank}_{i}},$
where $Q$ is the number of queries and $\mathrm{rank}$ is the position of the ground truth answer document among ranked candidates.

\textbf{Clone Detection}.
In the clone detection task, the problem is to retrieve semantically similar codes given a code as the query. To train and test models for clone detection, we use the \textbf{POJ-104} dataset introduced by
\citet{POJ}. It comes from a pedagogical programming open judge (OJ) system that automatically judges the validity of submitted source code for specific problems by running the code. The \poj dataset consists of 104 problems and includes 500 student-written C/C++ programs for each problem. The clone detection here is, given a program's source code, to retrieve other programs that solve the same problem.
The problems are grouped into three sets with 64/16/24 problems for training, validation, and testing respectively.
The default metric for the \poj dataset is Mean Average Precision (MAP), where the average precision (AP) is defined as
$\mathrm{AP} = \sum_{i=1}^{100} (R_i - R_{i-1})\cdot P_i,$
where $R_i$ and $P_i$ are the precision and recall at threshold $i$, i.e., computed taking into account only top $i$ items from the candidate list. MAP is the mean AP over all queries. It is important to mention that for \poj the maximal possible $i$ is 499 since there are only 500 candidates in total.

\subsection{\xcd Dataset}
\label{sec:xcd}
In previous works, multilingual abilities of code language models were not sufficiently investigated. To fill this gap, we introduce a new multilingual clone detection/code retrieval dataset \xcd. The dataset is evaluated in two different settings: clone detection approach similar to the BUCC dataset~\cite{xu-etal-2018-unsupervised-cross}, \textit{retrieval style} clone detection similar to \poj~\cite{POJ}, and a \emph{hybrid} approach.
We use the \emph{CodeForces} submissions dump as the data source,
randomly choosing 110 problems and 100 accepted solutions in one language for each problem. We chose 5 languages, namely Python, Java, C\#, C++, C,
getting
\numprint{55000} code snippets in total.

\textbf{Full comparison evaluation setup}.
Here the task is interpreted as binary classification, similar to
\citet{xu-etal-2018-unsupervised-cross}. We evaluate a model on each pair from test set, which gives rise to $n^2$ comparisons. Each pair could be either positive, if the pair consists of solutions for the same problem, or negative otherwise. We use the classic $F_1$ measure for evaluation in this setup:
$F_{1}=\frac{2\cdot\mathrm{Precision} \cdot \mathrm{Recall}}{\mathrm{Precision} + \mathrm{Recall}}.$

\textbf{Retrieval style evaluation setup}.
For retrieval style evaluation, we follow the design of \poj.
The task aims to retrieve 100 snippets per language solving the same problem from the test set. We evaluate ranking on \numprint{11000} positive snippets, i.e., the model should rank \numprint{11000} documents, bringing on top the snippets solving the same problem. Similar to
\citet{POJ}, we use mean average precision on the top 100 results (MAP@100) for evaluation.

\textbf{Hybrid evaluation setup}.
We also propose a setup where we evaluate the models not only on positive snippets but on all snippets in the same language. This task is harder and more similar to \advtest. As the metric we chose MRR@R, following
\citet{lu_codexglue_2021}.

\textbf{Cross-lingual evaluation}.
In addition to the \emph{multilingual} setting, i.e., evaluation in the set of solutions in a single language, we define a more complex \emph{cross-lingual} setting designed to measure cross-lingual code understanding. To achieve this, we extend the three setups to all languages at once, extending the sets of both relevant and irrelevant snippets to all programming languages.

\subsection{Additional Labeling}\label{sec:addlabel}
In addition to the solution status (``Accepted'' or not), we also mined error statuses for the solutions since the platforms used for problem solving often provide them. In total, we mined more than 97 million code snippets in more than 10 programming languages.
The \emph{CodeForces} platform can return 15 types of verdicts for a submitted solution. We split the verdicts into 4 groups: \emph{Defect} (code has a defect), \emph{Skip} (code cannot be judged), \emph{Accepted} (no defects detected), and \emph{Wrong} (code that fails some tests or constraints). Below we describe and classify some of the most common verdicts:
\begin{inparaenum}[(1)]
\item \emph{Memory limit exceeded}: the program tries to consume more memory than allowed (\emph{Wrong});
\item \emph{Time limit exceeded}: the program has not terminated in allotted time (\emph{Wrong});
\item \emph{Runtime error}: the program terminated with a non-zero return code (e.g., due to division by zero, stack overflow etc.) (\emph{Defect});
\item \emph{Wrong answer} on some tests (\emph{Wrong});
\item \emph{Idleness limit exceeded}: the program did not use the CPU for a considerable time (\emph{Defect});
\item \emph{Denial of judgement}: the solution was impossible to run due to a judging error or an error in the program, e.g., using extra large arrays (\emph{Defect});
\item \emph{Rejected}: the program does not pass tests for an unclear reason (\emph{Skip});
\item \emph{Accepted}: the program passed all tests.
\end{inparaenum}

\section{Method}\label{sec:method}

\begin{figure}[!t]
\centering
\includegraphics[width=\linewidth]{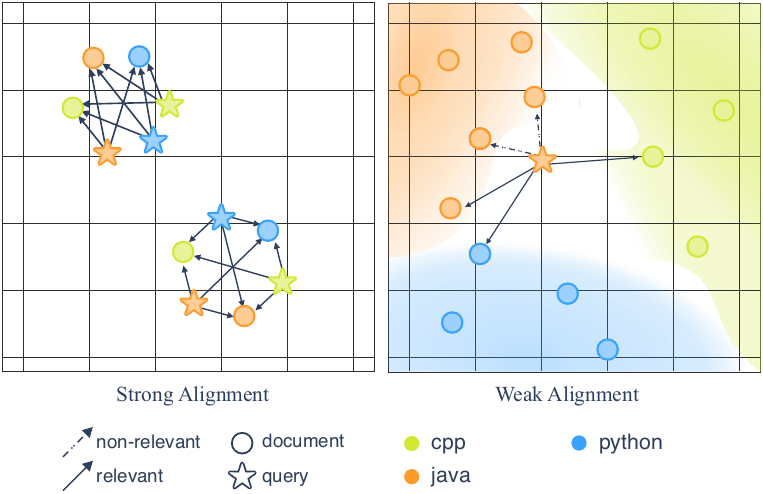}
\caption{Strong and weak cross-lingual alignment.}\label{fig:align}
\end{figure}

In this section, we introduce our pre-training approach \cct (Cross-Consistency Training). Its goal is to robustly learn the embedding space of code snippets and create a \emph{strong} alignment between snippets solving the same problems across programming languages. The difference between strong and weak alignment is illustrated in Fig.~\ref{fig:align}: in a weakly aligned embedding space, the nearest neighbor might be a semantically similar snippet from a different language but generally most neighbors are in the same language, while in a strongly aligned space the similarity is purely semantic and does not care about the language at all.

\def\ll{\mathcal{L}}
\def\lerr{\ll_{\mathrm{err}}}
\def\llm{\ll_{\mathrm{LM}}}
\def\lxcd{\ll_{\mathrm{XCD}}}
\def\hz{\hat{\z}}
\def\cQ{\mathcal{Q}}
\def\cW{\mathcal{W}}
\def\wxcd{\cW_{\mathrm{XCD}}}
\def\wlm{\cW_{\mathrm{LM}}}
\def\wbce{\cW_{\mathrm{BCE}}}
\def\lcl{\ll_{\mathrm{CL}}}

To achieve strong alignment, we employ a contrastive learning objective $\lxcd$: for a randomly code snippet, we train the vector representations of the source code tokens in such a way that their aggregation, for example, averaging or last token, is closer to the source code, which solves the same problem regardless of the programming language.
This ensures that the embeddings of the source code differentiates between related snippets and random or similar but different (hard negative) snippets effectively.

\textbf{Noise-contrastive estimation and losses}.
To learn a language-agnostic cross-lingual representation space, we propose a training procedure based on noise contrastive estimation (NCE).
Let $\mathcal{X}$ and $\mathcal{Z}$ be some finite sets and $\sth: \mathcal{X} \times \mathcal{Z} \rightarrow \RR$ be a relevance score function differentiable in $\btheta \in \RR^d$. The goal is to learn $\btheta$ such that
the classifier $\x \mapsto \argmax_{\z \in \mathcal{Z}} \sth(\x, \z)$ has the optimal expected loss. This leads to conditional density estimation: for every $\x \in \mathcal{X}$
\begin{equation}\label{eq:cond_prob}
\pcth{\z}{\x} = \frac{\myexp{\sth(\x,\z)}}{\sum_{\z^- \in \mathcal{Z}} \myexp{\sth(\x,\z^-)}}
\end{equation}
with $\btheta^* = {\arg\,\min\limits_{\btheta}}\,\EEEE{\x,\z}{-\log\pcth{\z}{\x}}$ being the optimum.
In practice, optimizing this objective directly is infeasible: if $\mathcal{Z}$ is large the normalization term in~\eqref{eq:cond_prob} is intractable.
Therefore, NCE uses subsampling, so~\eqref{eq:cond_prob} becomes
\begin{equation}\label{eq:pi}
\picth{\z}{\x} = \frac{\myexp{\sth(\x,\z)}}{\sum_{\z^- \in \mathcal{B}_{\x,\z}} \myexp{\sth(\x,\z^-)} +\myexp{\sth(\x,\z)}},
\end{equation}
where $\mathcal{B}_{\x,\z} =\{\z_1^-,\z_2^-,\ldots,\z_n^-\}$ is a set of \emph{negatives} sampled from $\mathcal{Z}$ that do not match the \emph{positive} answer $\z^+$ for this $\x$.
NCE also often uses objectives similar to~\eqref{eq:pi} but with $\picth{\hz}{\z}$ where $\z$ and $\hz$ come from the same space, and the objective corresponds to some similarity function.

\textbf{Cross-lingual objective}. Contrastive learning frequently employs pretext tasks to learn data representations without the need for labeled examples. In the context of learning from a multilingual set of documents, a possible pretext task would be to train a network to differentiate between documents with similar content but written in different languages (positive pairs) and those with dissimilar content (negative pairs). This leads to the loss function
Training with this task leads to learning a representation space with strong alignment properties, but the IR ability of the representations is mediocre.

\begin{equation}\label{eq:nce_xcd}
\lxcd(\btheta) = \EEEE{{(\hz,\z)} \sim \wxcd}{-\log \picth{\hz}{\z}},
\end{equation}
where $\wxcd$ is a distribution on the set of pairs of submissions in different programming languages from the \xcd dataset (Section~\ref{sec:data}) that shows if the submissions are solving the same problem or not.

\textbf{Hard negative mining}.
Previous works on contrastive learning show the importance of training on hard negative samples~\cite{Qu2021RocketQAAO, izacard2020distilling}. They used iterative training to get hard negatives, but our data already contains strong negative examples as preliminary solutions from the same users that solve the same problems but fail some tests (that is why a user would submit an updated solution to get the ``Accepted'' verdict). Thus, we mine hard negative examples as failed solutions from the same user; if there are none we use failed solutions from random users, and only if there are none (e.g., for an unpopular problem) we use a random submission for a random problem.

\section{Experiments}\label{sec:eval}

In this section, we describe the details about data pre-training and our \cct pipeline for multilingual clone detection and code search tasks.

\textbf{Pretraining}.
We train two models, one is encoder-based, which is initialized with pretrained \gcbertb model~\cite{guo_graphcodebert_2021}; we call the resulting model $\cctlm_{\mathrm{enc}}$. Another one is decoder-based, which is initilized with a pretrained DeepSeek-Coder-1.3B model~\cite{deepseekcoder}; we call the resulting model $\cctlm_{\mathrm{dec}}$.
Similarity scores are calculated based on dot products of the last token vector representations, but we also researched using various types of poolings and allowing bidirectional attention.

\textbf{Hyperparameters}. We use the AdamW optimizer with learning rate 5e-5, weight decay 0.01, and linear learning rate decay. We use gradient accumulation for pretraining with an effective batch size of 2000.

\begin{table}[!t]
\centering\setlength{\tabcolsep}{3pt}
\resizebox{\linewidth}{!}{
\begin{tabular}{l|c|c}
\toprule
& \textbf{Clone}  & \textbf{Code} \\
& \textbf{detection} & \textbf{search} \\
& (\textbf{MAP}) & (\textbf{MRR})  \\ \midrule
\multicolumn{3}{c}{\textbf{Endcoder-only}} \\ \midrule
RoBERTa-base \cite{RoBERTa} & 76.67  & 18.33 \\
\cbert \cite{feng_codebert_2020} & 82.67 & 27.19 \\
SynCoBERT \cite{wang_syncobert_2021} & 88.24 &  38.10 \\
CodeRoBERTa & --- & 42.35 \\
\gcbert \cite{guo_graphcodebert_2021} & 85.16 &  --- \\
CasCode~\cite{gotmare2021cascaded} &  --- &  43.98 \\
\citet{AddLeak}  &  91.34 &  --- \\
$\cctlm_{\mathrm{enc}}$  & \textbf{96.73} & \textbf{47.18} \\
\cmidrule{1-3}
\multicolumn{3}{c}{\textbf{Decoder-only}} \\ \midrule
CodeGen \cite{codegen}  & 89.68 & --- \\
CodeGPT \cite{codegpt}  & 87.96 & --- \\
SantaCoder \cite{santacoder}  & 83.98 & --- \\
Phi-1 \cite{phi-1}  & 92.72 & --- \\
$\cctlm_{\mathrm{dec}}$  & \textbf{95.50} & --- \\
\bottomrule
\end{tabular}
}
\caption{Results on code clone detection on the \poj dataset and code search on the \advtest dataset.}
\label{tab:results}
\end{table}

\textbf{Monolingual Results}.
Tab.~\ref{tab:results} presents the results of \cctlm models compared to existing approaches, showing that \cctlm outperform all previous models by a large margin in this monolingual setting. Thus, strong alignment enforced by \cct pretraining is not only helpful for multilingual transfer but also improves the latent space structure in general. It is important to mention, that \cct pretraining works for both encoder- and decoder-based models, improving the results.

\begin{table*}[!t]
\centering\setlength{\tabcolsep}{6pt}
\resizebox{\textwidth}{!}{
\begin{tabular}{l|ccccccccc|c}
\toprule
&  \textbf{Python} &  \textbf{Java} &  \textbf{C\#} &  \textbf{Ruby} &  \textbf{JS} &  \textbf{Haskell} &  \textbf{PHP} &  \textbf{OCaml} &  \textbf{Perl} &  \textbf{Avg} \\ \midrule\midrule
\multicolumn{11}{c}{{\textbf{Multilingual setting}}} \\\midrule
\multicolumn{11}{c}{\textbf{Full Comparison, $F_1$ measure}} \\\midrule
\gcbertb &  0.02 &  0.05 &  0.00 &  0.04 & 0.00 &  0.02 &  0.01 &  0.03 & 0.01 &  0.02  \\
\gcbertbp &  0.04 &  0.00 & 0.01 & 0.06 &  0.07 & 0.08 &  0.06 & 0.06 & 0.06  & 0.05 \\
$\cctlm_{\mathrm{enc}}$ &  \textbf{22.24} & \textbf{18.39} &   \textbf{17.33} &    \textbf{23.33} &    \textbf{10.46} &    \textbf{17.64} &    \textbf{21.43} &\textbf{17.01}& \textbf{16.40} &    \textbf{18.24}   \\\midrule
\multicolumn{11}{c}{\textbf{Retrieval Style, MAP@100}}   \\\midrule
BM25 &  0.00 &  0.00 &  0.00 &  0.00 &  0.00 &  0.00 &  0.00 &  0.00 &  0.00 &  0.00 \\
\gcbertb &  7.21 &  9.25 &  1.33 &  4.28 &  1.59 &  5.78 &  6.08 &  2.90 & 10.37 &  5.42  \\
\gcbertbp &  30.12 &  24.63 &  23.54 &  32.78 &  36.64 & 24.45 &  37.21 & 33.94 &  45.33 & 32.07 \\
$\cctlm_{\mathrm{enc}}$ &    \textbf{87.42} & \textbf{55.99} &    \textbf{65.35} &    \textbf{72.12} &    \textbf{74.32} &    \textbf{81.05} &    \textbf{83.21} &\textbf{71.53}& \textbf{71.89} &    \textbf{73.65}   \\\midrule
\multicolumn{11}{c}{\textbf{Hybrid, MRR@20}}   \\\midrule
BM25 &  0.00 &  0.00 &  0.00 &  0.00 &  0.00 &  0.00 &  0.00 &  0.00 &  0.00 &  0.00 \\
\gcbertb &  2.08 &  5.42 &  0.22 &  2.59 & 0.80 &  1.99 &  2.90 &  1.40 & 5.23 &  2.51  \\
\gcbertbp &  27.10 &  20.04 &  19.44 &  30.98 &  28.37 & 19.70 &  32.89 & 30.08 &  39.98 & 27.62 \\
$\cctlm_{\mathrm{enc}}$ &    \textbf{74.97} & \textbf{62.08} &   \textbf{58.77} &    \textbf{80.60} &    \textbf{74.56} &    \textbf{62.27} &    \textbf{81.21} &\textbf{72.64}& \textbf{79.16} &    \textbf{71.80}   \\\midrule
\midrule
\multicolumn{11}{c}{{\textbf{Cross-lingual setting}}} \\\midrule
\multicolumn{11}{c}{\textbf{Full Comparison, $F_1$ measure}}   \\\midrule
\gcbertb &  0.01 &  0.01 &  0.00 &  0.01 &  0.00 &  0.01 &  0.01 &  0.01 & 0.01 &  0.01  \\
\gcbertbp &  0.01 &  0.00 & 0.01 & 0.01 &  0.01 & 0.01 &  0.01 & 0.01 & 0.01  & 0.01 \\
$\cctlm_{\mathrm{enc}}$ &  \textbf{8.92} & \textbf{9.46} &   \textbf{4.78} &    \textbf{6.01} &    \textbf{7.33} &    \textbf{5.82} &    \textbf{6.47} &\textbf{5.33}& \textbf{3.56} &    \textbf{6.40}  \\\midrule
\multicolumn{11}{c}{\textbf{Retrieval Style, MAP@100}}   \\\midrule
BM25 &  0.00 &  0.00 &  0.00 &  0.00 &  0.00 &  0.00 &  0.00 &  0.00 &  0.00 &  0.00 \\
\gcbertb &  3.18 &  5.24 &  0.23 &  1.77 &  1.15 &  3.38 &  3.12 &  1.90 & 16.27 &  4.02  \\
\gcbertbp &  12.83 &  14.75 &  9.33 &  12.78 &  17.16 & 15.94 &  19.53 & 16.01 &  23.88 & 15.80 \\
$\cctlm_{\mathrm{enc}}$    &    \textbf{44.82} & \textbf{20.34} &    \textbf{23.33} &    \textbf{35.01} &    \textbf{32.57} &    \textbf{40.07} &    \textbf{43.36} &\textbf{36.66}& \textbf{37.80} &    \textbf{34.88}    \\\midrule
\multicolumn{11}{c}{\textbf{Hybrid, MRR@20}}   \\\midrule
BM25 &  0.00 &  0.00 &  0.00 &  0.00 &  0.00 &  0.00 &  0.00 &  0.00 &  0.00 &  0.00 \\
\gcbertb &  1.24 &  2.42 &  0.34 &  1.28 & 0.82 &  0.93 &  1.43 &  0.76 & 2.15 &  1.26  \\
\gcbertbp &  20.12 &  13.08 &  10.37 &  17.28 &  12.62 & 19.70 &  14.31 & 18.08 &  18.33 & 15.98 \\
$\cctlm_{\mathrm{enc}}$ &    \textbf{30.83} & \textbf{22.77} &   \textbf{19.32} &    \textbf{32.66} &    \textbf{31.64} &    \textbf{20.80} &    \textbf{31.59} &\textbf{40.42}& \textbf{39.40} &    \textbf{29.93}   \\\bottomrule
\end{tabular}
}
\caption{Multilingual clone detection in two evaluation setups on the \xcd dataset.}\label{tbl:res}
\end{table*}
\subsection{Multi- and Cross-lingual Evaluation}
For these types of evaluation on \xcd we use several setups described in Sec.~\ref{sec:xcd}. Since these setups are computationally intensive we work  only with encoder-based models.

\textbf{Multilingual Results}.
Results in the multilingual setting on the proposed \xcd dataset are shown in the top half of Tab.~\ref{tbl:res}.
In the full comparison setup,
it is interesting that knowledge transfer from the \poj dataset also does not help, and the metrics remain very low. However, \cctlm shows much better results, obviously due to the multilingual design of \cct pretraining. We do not evaluate BM25 in this setup since it is not supposed to compare two documents.

For retrieval style evaluation, the results
also show $\cctlm_{enc}$ strongly outperforming all baselines and actually providing a viable solution for the problem while \gcbert does not yield reasonable results in any programming language. We note that BM25 (a strong baseline for natural language IR tasks) does not work for clone detection, which is natural since it relies on identical tokens which could be scarce even between code snippets solving the same problem.

Results for the hybrid evaluation setup
show the same picture: BM25 still does not work, code language models can transfer knowledge across different solutions to some extent, and the improvement in metrics here between a fully unsupervised method and training on \poj clone detection is noticeably greater. Still, $\cctlm_{enc}$ far exceeds every other method here and generally sets a new baseline for multilingual code-related tasks.

\textbf{Cross-lingual Results}.
Our results in this setting are presented in the bottom half of Tab.~\ref{tbl:res}. All conclusions derived for the multilingual case (above) apply here too, but in comparison to the multilingual setting, cross-lingual tasks are significantly harder and all values are lower. We suggest that the difference in the results across programming languages could be caused by the imbalance in the pretraining dataset.

\section{Analysis}\label{sec:analysis}

\begin{table}[t!]
\centering\setlength{\tabcolsep}{2pt}
\resizebox{\linewidth}{!}{
\begin{tabular}{l|ccccc|c}
\toprule
&  \textbf{Java} &  \textbf{Ruby} & \textbf{PHP}  & \textbf{Go} & \textbf{JS} &  \textbf{Avg} \\ \midrule
\cbertb & 46.37 & 50.65 & 37.83 & 50.65 & 50.48 & 47.19  \\
\gcbertb & 47.33 & 59.95 & 37.47 & 60.28 & \textbf{52.04} & 51.41 \\
$\cctlm_{enc}$ & \textbf{48.71} & \textbf{62.25} & \textbf{42.78} & \textbf{61.44} & 51.06 & \textbf{53.24}   \\\bottomrule
\end{tabular}
}
\caption{Zero-shot retrieval; $F_1$ score, \csn.}
\label{tab:zeroshot}\vspace{.2cm}

\resizebox{\linewidth}{!}{
\begin{tabular}{l|c|c|c}
\toprule
& \textbf{Clone}  & \textbf{Code} & \textbf{Defect} \\
& \textbf{Detection} & \textbf{Search} & \textbf{detection} \\
\gcbert & (\textbf{MAP}) & (\textbf{MRR}) & (\textbf{Acc}) \\ \midrule
Base & 85.16 &  45.80 & 62.51 \\
Base + $\lxcd$ & 95.92 & 29.93 & 61.05 \\
Base + $\lxcd$ + $\llm$ & 95.67 & 47.18 & 63.68 \\
Base +  $\lxcd$ + $\llm$ & 96.03 & 45.22 & 64.91 \\
Base +  $\lxcd$ + $\llm$ + SL & 96.46 & 47.33 & - \\
Base +  $\lxcd$ + $\llm$ + $\lerr$ + SL & \textbf{96.73}  & \textbf{47.57} & \textbf{65.58} \\
\bottomrule  \end{tabular}
}
\caption{\gcbert variations: clone detection on \poj, code search on \advtest, defect detection on \emph{Devign}; SL denotes the size limit.}
\label{tab:analysis}
\caption{A comparison of DeepSeek-Coder 1.3b variations: clone detection on \poj, code search on \advtest
}
\label{tab:analysis}
\end{table}

\textbf{Zero-shot Results}.
We investigated zero-shot transfer from Python to Java, Ruby, PHP, Go, and JavaScript on the \csn dataset for previously introduced code language models and our \cctlm.
The zero-shot results are presented in Table~\ref{tab:zeroshot}.
As evidence for the power of pretrained language models, we see that existing approaches show rather good results even though they have not been trained on the retrieval task. By leveraging its multilingual ability, \cctlm improves over the baselines in the zero-shot setup for all languages except \emph{JavaScript} (JS).

\begin{figure}[t!]
\centering
\begin{subfigure}[b]{.9\linewidth}
\centering
\includegraphics[width=\linewidth]{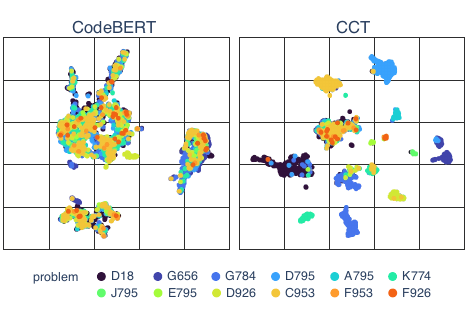}
\caption{Projected embeddings of 12 coding problems.}
\label{fig:wiki_align}
\end{subfigure}
\hfill
\begin{subfigure}[b]{.9\linewidth}
\centering
\includegraphics[width=\linewidth]{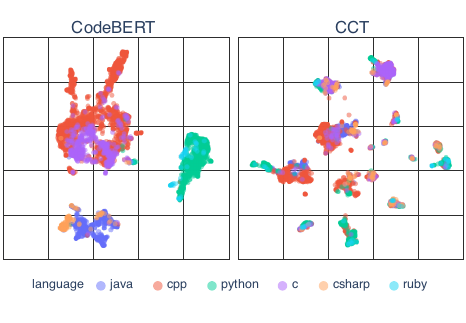}
\caption{The same embeddings by programming language.}
\label{fig:same_lang}
\end{subfigure}
\caption{Sample multilingual embeddings.}
\label{fig:wiki_abs_analysis}
\end{figure}

\textbf{Latent space structure}.
Figure~\ref{fig:align} showed an abstract representation of the basic \cct idea of semantically aligned language-agnostic embedding space. Figure~\ref{fig:wiki_abs_analysis} turns this theory into practice with projections of actual embeddings for sample code snippets before and after \cct training. The snippets represent solutions for 12 sample tasks in six programming languages. We see that after \cct, representations of code snippets are not aligned by language but rather by problem (Fig.~\ref{fig:same_lang}), while their alignment had been language-dependent before \cct (Fig.~\ref{fig:wiki_align}).

This illustrates that \cct training significantly improves the multilingual latent space for code snippets, making it semantic and language-agnostic.

\textbf{Ablation Study}.
In this section, we study the effects of various parts of \cct. Table~\ref{tab:analysis} shows the results of several DeepSeek-Coder-based models on clone detection, code search tasks.
We compare the DeepSeek-Coder base model with different pretraining poolings and attention types.

\section{Conclusion}\label{sec:concl}

Understanding semantic similarity is an important aspect of language processing that opens up ways to solve many different tasks, for both natural and programming languages. In this work, we have presented a new method \cctlm that improves this ability via a novel \cct pretraining approach and demonstrated its viability for both clone detection and code search. We have formulated a novel task of multilingual clone detection and presented the \xcd dataset for multilingual source code analysis, formalized in two evaluation setups.
The proposed \cctlm models (both encoder- and decoder-based) outperformed strong baselines in clone detection and code search task. $\cctlm_{enc}$ exceed other models in all the setups for multi- and cross-lingual evaluation, proving that \cct pretraining provides better semantic similarity understanding for a language model.
We hope that our method will be helpful in other source code processing tasks that we have left as future work. Moreover, we believe that modifications of our approach can be useful for NLP and perhaps other fields of machine learning.

\section{Limitations}\label{sec:limits}

We have studied several programming languages, including Python and Java, in our XCD setup; although all our methods seem to be language-agnostic, a further study for other languages would be interesting, especially since all considered languages are interpreted rather than compiled (like C/C++).
Many inputs exceed 512 tokens; we used standard truncation for evaluation (taking into consideration only the beginning of the code), which may be suboptimal, and more suitable input representations could be found.
We expect our model to improve with training on long documents.
We also suppose that the model would benefit from increasing the batch size by using more powerful hardware with more memory.
Note also that while \cctlm significantly improved state of the art in clone detection and code search.

\bibliography{anthology, custom}

\end{document}